\documentclass[10pt,twocolumn,letterpaper]{article}

\usepackage{wacv}
\usepackage{times}
\usepackage{epsfig}
\usepackage{graphicx}
\usepackage{amsmath}
\usepackage{amssymb}
\usepackage{booktabs}
\usepackage[accsupp]{axessibility} % Improves PDF readability for those with disabilities.

\newcommand*{\textcal}[1]{%
  \textit{\fontfamily{qzc}\selectfont#1}%
}

\def\wacvPaperID{1380} % Enter the WACV Paper ID here

\wacvfinalcopy % *** Uncomment this line for the final submission

\ifwacvfinal
\usepackage[breaklinks=true,bookmarks=false]{hyperref}
\else
\usepackage[pagebackref=true,breaklinks=true,colorlinks,bookmarks=false]{hyperref}
\fi

\begin{document}

%%%%%%%%% TITLE
\title{CameraPose: Weakly-Supervised Monocular 3D Human Pose Estimation\\ by Leveraging In-the-wild 2D Annotations}

\author{Cheng-Yen Yang$^{1*}$, Jiajia Luo$^{2}$, Lu Xia$^{2}$, Yuyin Sun$^{2}$, Nan Qiao$^{2}$, Ke Zhang$^{2}$,\\
Zhongyu Jiang$^{1}$, Jenq-Neng Hwang$^{1}$, Cheng-Hao Kuo$^{2}$\\
$^{1}$ Department of Electrical and Computer Engineering, University of Washington, WA, USA \\ 
$^{2}$ Amazon Lab126, USA \\
{\tt\small \{cycyang,zyjiang,hwang\}@uw.edu}, \\ \tt\small\{lujiajia,luxial,yuyinsun,kezha,qiaonan,chkuo\}@amazon.com}
% For a paper whose authors are all at the same institution,
% omit the following lines up until the closing ``}''.
% Additional authors and addresses can be added with ``\and'',
% just like the second author.
% To save space, use either the email address or home page, not both
% \and
% Second Author\\
% Institution2\\
% First line of institution2 address\\
% {\tt\small secondauthor@i2.org}

\maketitle
\thispagestyle{empty}

%%%%%%%%% ABSTRACT
\begin{abstract}
     To improve the generalization of 3D human pose estimators, many existing deep learning based models focus on adding different augmentations to training poses. However, data augmentation techniques are limited to the "seen" pose combinations and hard to infer poses with rare "unseen" joint positions. To address this problem, we present CameraPose, a weakly-supervised framework for 3D human pose estimation from a single image, which can not only be applied on 2D-3D pose pairs but also on 2D alone annotations. By adding a camera parameter branch, any in-the-wild 2D annotations can be fed into our pipeline to boost the training diversity and the 3D poses can be implicitly learned by reprojecting back to 2D. Moreover, CameraPose introduces a refinement network module with confidence-guided loss to further improve the quality of noisy 2D keypoints extracted by 2D pose estimators. Experimental results demonstrate that the CameraPose brings in clear improvements on cross-scenario datasets. Notably, it outperforms the baseline method by 3mm on the most challenging dataset 3DPW. In addition, by combining our proposed refinement network module with existing 3D pose estimators, their performance can be improved in cross-scenario evaluation.   
\end{abstract}

\let\thefootnote\relax\footnotetext{* This work was mostly done when Cheng-Yen Yang was an intern at Amazon Lab126.}

\section{Introduction}\label{sec:intro}

Human pose estimation (HPE) is a task to predict the configuration of a particular set of human body parts from some visual input such as images or videos. Depending on the output format, it can be further divided into 2D and 3D HPE, respectively. Different from the 2D HPE that predicts the human keypoints with $x,y$ coordinates, the 3D HPE regresses $x,y,z$ which can be more helpful to solve difficult tasks, such as action and motion prediction\cite{application_motion_scene, application_motion_corona}, posture and gesture recognition \cite{ application_action_recognition_lienet, application_action_recognition}, augmented reality and virtual reality \cite{application_VRAR_multiperson, application_VRAR_hand}, healthcare \cite{application_health_inbed, application_health_dementia}. Although deep learning based methods have boosted the performance of 3D HPE \cite{martinez_2017_3dbaseline, 3dhp, DBLP:journals/corr/abs-1805-04092, videopose, Zhou_2017_ICCV}, the error will typically increase to around two times from Human3.6M \cite{h36m_pami} to 3DHP \cite{3dhp} for cross-dataset scenario due to the poor model generalization \cite{poseaug}.

\begin{figure}[t]
  \centering
  \includegraphics[width=1.\linewidth]{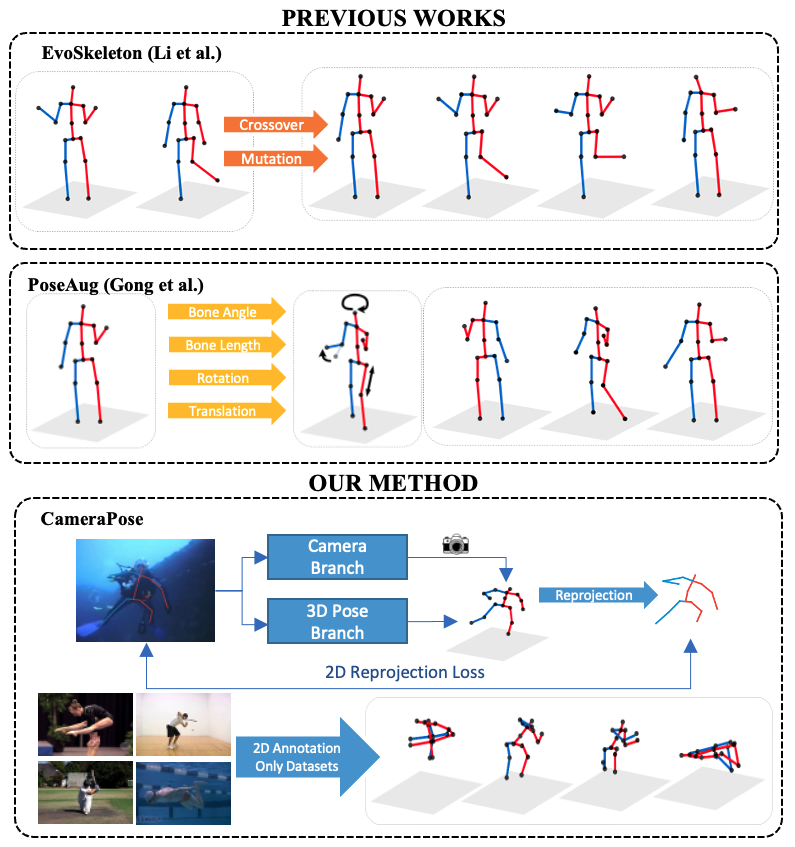}
  \caption{Training data expansion overview. Data augmentation on existing 2D poses can improve the diversity of training to some extend. By taking advantage of in-the-wild 2D annotations, more rare but challenging poses can be utilized to further improve the model generalization.}
  \label{fig:evo_poseaug_ours}
\end{figure}

Recent works argue that poor model generalization can be mitigated by increasing the variance in training data. Therefore, many augmentation-related algorithms have been proposed to improve the 3D HPE accuracy. However, no matter it is image-based augmentation \cite{single_shot_data_augmentation, mocap_guided_data_augmentation}, synthetic-based augmentation~\cite{synthetic_data_augmentation, Vyas_2021_CVPR}, predefined transformation~\cite{evoskeleton} or GAN-based augmentation~\cite{poseaug}, the variances added to the training data is still limited to the original 2D-3D pair. Figure~\ref{fig:evo_poseaug_ours} shows examples of augmented 2D-3D pairs with different algorithms. We can observe that the generated new pair 2D-3D cannot provide pose changes (lying to sitting etc.). Due to the limitation in the training data, the scenes or scenario are still relatively simple to the in-the-wild environment, which hinder the real-world application of these algorithms. 

Different from the existing methods that rely on data augmentation for training data expansion, we proposed a novel weakly-supervised framework, CameraPose, to improve model generalization on 3D HPE by taking advantage of plentiful 2D annotations. Compared to the expensive 3D annotations, 2D annotations are less expensive, and many challenging 2D datasets \cite{ dataset_mpii, lsp,  coco} containing rich actions, poses, and scenes are available in the literature. The proposed CameraPose network can combine any existing 2D or 3D datasets in a single framework by adding a camera parameter estimation branch. Our approach also integrated the GAN-base pose augmentation framework to improve the training data diversity and ensure the camera branch's generalization.

\begin{figure}[t]
  \centering
  \includegraphics[width=0.82\linewidth]{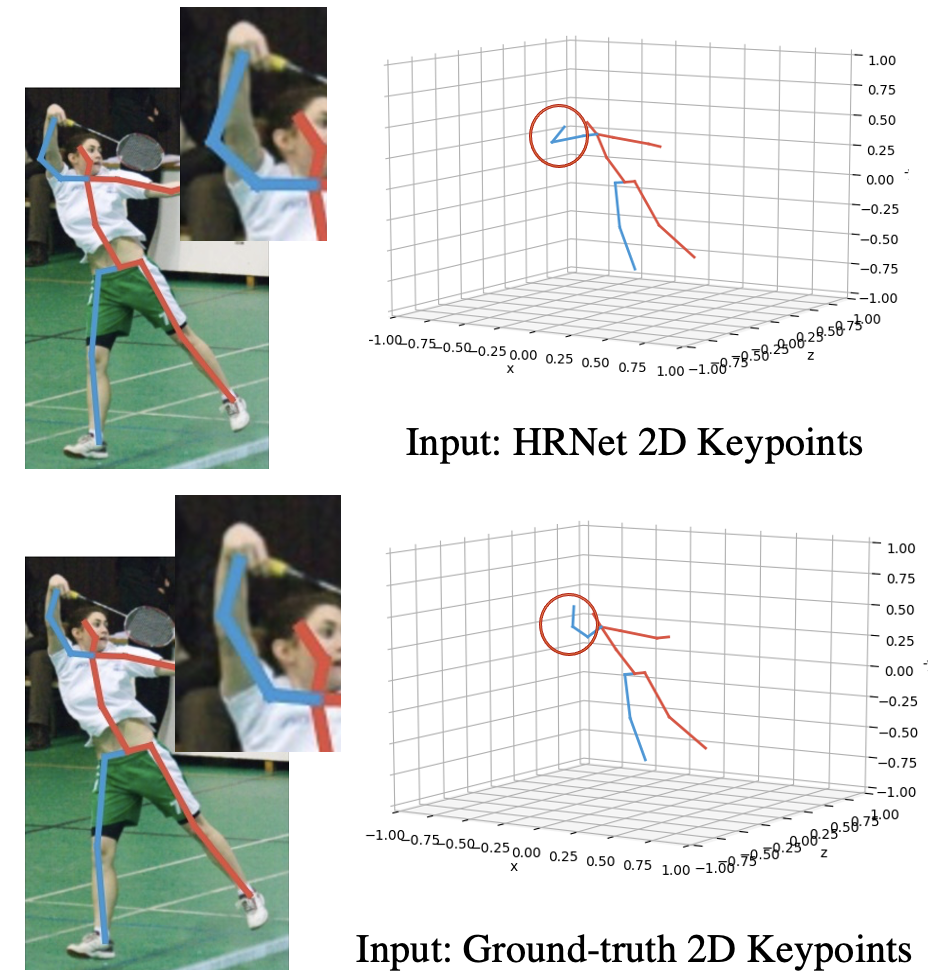}

  \caption{Example of feeding different source of 2D joints prediction into the same 3D lifting network. Due to the inaccurate right elbow prediction from the HRNet\cite{hrnet}, the errors from the same keypoint will be enlarged in the 3D poses.}
  \label{fig:difference_hr_gt}
  \vspace{-1em}
\end{figure}

Existing 3D HPE networks usually directly use 2D keypoints from some pre-trained detectors as input to train 3D joints. However, inferred 2D keypoints will lead to the situation illustrated in Fig.\ref{fig:difference_hr_gt}. The errors from the 2D joints estimation step will generate 3D prediction errors on some keypoints. In addition, augmentation on inaccurate 2D keypoints will further enlarge the errors in 3D joints.  As shown in Table ~\ref{table:gt}, the ground-truth inputs significantly boosted the accuracy in all testing cases with different pose estimators. Therefore, it is necessary to improve the 2D keypoints before feeding them into our 3D estimator network. To mitigate the error in 2D input, we propose to incorporate a refinement network that aims to infer better 2D joints based on the positions and confidence scores of detected 2D joints.

\begin{table}[t]
\centering
\caption{MPJPE on Human3.6M using different source of 2D keypoints source: HRNet and ground-truth.}
\begin{tabular}{lcc}
\hline
\multicolumn{1}{c}{3D Pose Estimator}   & \multicolumn{2}{c}{\begin{tabular}[c]{@{}c@{}}Human3.6M\\ (MPJPE)\end{tabular}} \\ \hline
\multicolumn{1}{c}{2D Keypoints Source} & HRNet                               & Ground-truth                              \\ \hline
Zhao et al.      \cite{GCN}                       & 57.5                                & 44.4                                      \\
Martinez et al.   \cite{martinez_2017_3dbaseline}                      & 53.0                                & 43.3                                      \\
Pavllo et al.     \cite{videopose}                     & 52.2                                & 41.8                                      \\ \hline
%Ours (CameraPose)                       & 54.2                                & 38.9                                      \\ \hline
\end{tabular}
\label{table:gt}
\end{table}

Our contributions are three-fold: 1)  We propose a camera parameter branch that will generate per-instance camera parameter inference so that any existing 2D keypoints datasets (without 3D labeling) can be utilized in model training. 2) We propose a Refinement Network to improve the accuracy of 2D joints, which can be helpful in the GAN-based augmentation stage, as well as the final 3D joints predictions. 3) We introduce the reprojection loss, confidence-guided refinement loss, together with the camera loss in the loss design to make the network differentiable. 

\begin{figure*}[t]
  \centering
  \includegraphics[width=1.0\linewidth]{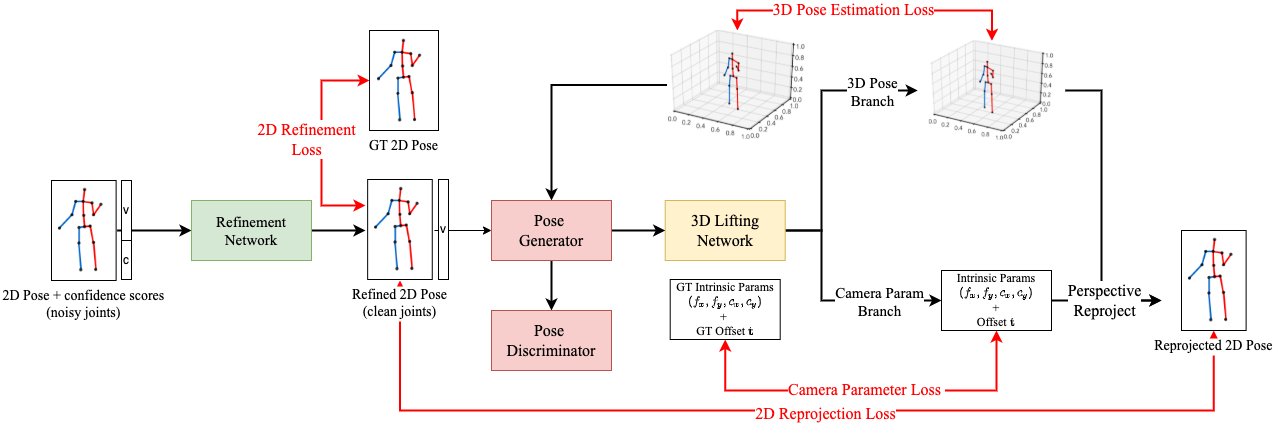}
  \caption{Overall framework of our proposed CameraPose. It consisted of three main parts: (1) RefineNet, (2) Pose Generator/Discriminator, and (3) Weakly-Supervised Reprojection Camera Branch. When trained with 2D-3D annotated datasets, all of the loss will be used while with 2D only datasets, only the 2D projection loss will be considered to update the weights.}
  \label{fig:framework}
\end{figure*}

\section{Related Works}\label{sec:related}

\noindent \textbf{Fully-Supervised 3D HPE.} There are a lot of papers and research that use the 2D-3D annotation pairs for a fully-supervised training manner. Tekin et al. \cite{DBLP:journals/corr/TekinKSLF16} directly regress the 3D human pose from a spatio-temporal volume of bounding boxes, and Martinez et al. \cite{martinez_2017_3dbaseline} regress the 3D human pose from a naive MLP using 2D keypoints as input and 3D keypoints as output. 

On similar datasets, these end-to-end methods often perform very well. Their capacity to generalize to different settings, on the other hand, is restricted. Many studies use cross dataset training or data augmentation to address this issue \cite{mocap_guided_data_augmentation, single_shot_data_augmentation, synthetic_data_augmentation, Vyas_2021_CVPR}. Most recently, Li et al. \cite{evoskeleton} directly augment 2D-3D pose pairs by randomly applying partial skeleton recombination and joint angle perturbation on source datasets. Then Gong et al. \cite{poseaug} used a generative-based model to manipulate the transformation of 3D ground-truth and then do the reprojection back to image space to get the corresponding 2D keypoints. This can be trained along with the 3D lifting network and some discriminators to ensure the augmented poses are realistic and increase the diversity of the training dataset. While effective, the major downside of all supervised approaches is that they do not generalize well to unseen poses. Therefore, their application to in-the-wild scenes is limited.

Some even use a portion amount of dataset to do the training for human pose estimation through methods like transfer learning \cite{3dhp, 2_transfer_learning_sim2real, 3dpw}. As they all try to mixed 2D pose from in-the-wild images and 3D poses from laboratory settings to learn the deep features through shared representation. These methods generalize better to unseen poses because they learn distributions of realistic 3D postures and their characteristics. They can recreate out-of-distribution positions to a degree, but they have trouble with entirely undetected poses.

\noindent \textbf{Weakly-Supervised 3D HPE.} Some approaches use unpaired 2D-3D annotations to get some 3D priors or basis to do the 3D human pose estimation from a monocular camera. Drover et al. \cite{3d_from_2d_alone} proposed a projection layer that randomly projects the predicted 3D poses back into 2D poses and then feeds into a discriminator. Chen et al. \cite{2_ws_reprojection_self_geo} introduced cycle consistency loss into \cite{3d_from_2d_alone} extending the training with a step of lifting the projected 2D pose once again into the 3D pose. Habibie et al. \cite{habibie} designed an architecture that comprises an encoding of explicit 2D and 3D features, and uses supervision by a separately learned projection model from the predicted 3D pose. Wandt et al. \cite{repnet} proposed RepNet to tackle the problem with reprojection constraints by using an adversarial-based method with a sub-network that can estimate the camera. However, we argue the gap between supervised algorithms and unsupervised algorithms can be large on some challenging datasets.

As for multi-view settings, Rochette et al. \cite{2_ws_reprojection_consisency} using multi-view consistency by moving the stereo reconstruction problem into the loss. Kocabas et al. \cite{2_ws_reprojection_multi_view_geometry} proposed another multi-view approach by applying epipolar geometry to predicted 2D pose under different views to construct the pseudo-ground-truth. Iqbal et al. \cite{2_ws_reprojection_2.5d} proposed a end-to-end learning framework adopting a 2.5D pose representation without any 3D annotations. Wandt et al. \cite{canonpose} then proposed a self-supervised method that requires no prior knowledge about the scene, 3D skeleton, or camera calibration and also introduced the 2D joint confidences into the 3D lifting pipeline. However, these algorithms are hard to be applied to single-view or in-the-wild predictions due to their multi-view pipeline design.

\noindent \textbf{HPE with Data Augmentation.} Data augmentation can help the model generalization ability by enlarging the training data \cite{mocap_guided_data_augmentation, single_shot_data_augmentation, synthetic_data_augmentation, Vyas_2021_CVPR}. Most recently, Li et al. \cite{evoskeleton} directly augment 2D-3D pose pairs by randomly applying partial skeleton recombination and joint angle perturbation on source datasets. Then Gong et al. \cite{poseaug} used a generative-based model to manipulate the transformation of 3D ground-truth then do the reprojection back to image space to get the corresponding 2D keypoints. This can be trained along with the 3D lifting network and some discriminators to ensure the augmented poses are realistic and increase the diversity of the training dataset.

\vspace{-1em}
\section{Proposed Method}\label{sec:method}

The CameraPose network consisted of three main parts: (1) Refinement Network, (2) Pose Generator/Discriminator, and (3) Weakly-Supervised Camera Parameter Branch. Figure \ref{fig:framework} summarizes our CameraPose architecture design.

Let $x\in \mathbb{R}^{2\times N_{J}}$ denotes the 2D keypoints and $\textbf{X}\in \mathbb{R}^{3\times N_{J}}$ denotes the corresponding 3D joint position in the camera coordinate system with $N_J$ represents the number of joints in the framework. Our proposed network will train on two different cases of datasets: (1) 2D-3D annotated dataset $\boldsymbol\phi = (x, \textbf{X})$ ,and (2) 2D annotations only dataset $\boldsymbol\phi' = (\textbf{x'}, -)$ by optimizing the following equation:

\begin{equation}
    \small
    \underset{\theta_{3D}, \theta_{ref}}{\text{min}}\  \mathcal{L}_{\boldsymbol\phi} \Big( P_{\theta_{3D}} \big( R_{\theta_{ref}}(x) \big), \boldsymbol\phi \Big )  + \mathcal{L}_{\boldsymbol\phi'} \Big( P_{\theta_{3D}} \big( R_{\theta_{ref}}(x') \big), \boldsymbol\phi' \Big) 
\end{equation}

\noindent where $\theta_{3D}$ and $\theta_{ref}$ represent the weights of our 3D lifting model and refinement network. Furthermore we extend the design of pose augmentor$\mathcal{A}$ to enlarge the 2D-3D annotated dataset with the augmented dataset $\mathcal{A}(\phi)=(x^*, \textbf{X}^*)$. Therefore our end-to-end optimization procedure will become:

\begin{equation}
    \underset{\theta_{3D}, \theta_{ref}}{\text{min}} \underset{\theta_{\textcal{A}}}{\text{max}}\  \mathcal{L}_{\boldsymbol\phi}\big( \boldsymbol\phi \cup \mathcal{A}(\boldsymbol\phi) \big) + \mathcal{L}_{\boldsymbol\phi'} \big(\boldsymbol\phi'\big).
\end{equation}

\begin{table}[h]
\caption{Mathematical notations used in the equations.}
\begin{tabular}{cc}
\hline
Notation            & Description                                         \\ \hline
$N_J$               & number of joints used                               \\
$N_S$               & number of samples in the batch                      \\ \hline
$\boldsymbol\phi$   & datasets with 2D-3D annotations                     \\
$\boldsymbol\phi'$  & datasets with 2D annotations only                   \\
$\boldsymbol\phi^*$ & datasets generated by the pose generator            \\ \hline
$(x,\textbf{X})$                 & ground-truth 2D-3D annotations from $\boldsymbol\phi$  \\
$(x',-)$                & ground-truth 2D annotations from $\boldsymbol\phi'$ \\
$(x^*,\textbf{X}^*)$               & augmented 2D-3D annotations from $\boldsymbol\phi*$    \\
$\hat{\textbf{X}}$  & predicted 3D poses from 3D lifting network                                \\ \hline
\end{tabular}
\end{table}

\subsection{Refinement Network}

%First, we want to address the finding that not all 2D keypoint detections are perfect; the accuracy of those joints will be highly influenced by occlusion or unseen pose for our 2d joint detector. As shown in Figure ~\ref{fig:difference_hr_gt}, the inferred 3D positions for right elbow has larger error than other keypoints. This is mainly due to the larger error in the corresponding 2D joint obtained from HRNet \cite{hrnet}. To further prove our statement, we compared the 3D HPE errors from groundtruth vs HRNet based on different 3D pose estimators. As shown in Table ~\ref{table:gt}, the groundtruth inputs significantly boosted the accuracy in all testing cases with different pose estimators. Therefore, it is necessary to improve the 2D keypoints before feeding them into our 3D estimator network.

%Figure ~\ref{fig:confidence} shows the heatmap of each detected keypoints, and the right elbow has the lowest confidence score compared to the rest keypoints. 

Instead of refining on the original noisy 2D keypoints, we utilize the confidence score combined with the 2D $(x,y)$ coordinates as input to the refinement network. We first normalize the coordinates of keypoints to $(-1, 1)$ with respect to the input image height and width. We also normalized the confidence scores to a comparable scale by Eq.~\ref{eq:score_norm}:

%Since most 2D estimator simply regress the heatmap and choose the coordinates with maximum value as its joint prediction, this last layer of the network will not always be a sigmoid or softmax to regulate the outputs due to various 2D detector architecture design. Therefore, with the 2D keypoints coordinates being normalized to $(-1,1)$ with respect to the input image height and width, we also want to ensure that our confidence scores of these joints are in similar scale. The confidence scores was normalized using:

\begin{equation}
    \textbf{c}'_{ij} = \frac{\textbf{c}_{ij}}{||\textbf{C}_i||_{1}}
\label{eq:score_norm}
\end{equation}

\noindent where $||\cdot||_1$ denotes for L1 norm and $\textbf{C}_i$ stands for the all the heatmaps in the $i$-th training sample while $\textbf{c}_{ij}$ stands for the maximum value (confidence score) on the $j$-th heatmap. The normalized confidence score will be used as the weight to compute the joint-wise mean-square error in Eq. \ref{eq:loss_refine}. 

The neural network architecture of our Refinement Network is a standard residual block consisting of fully connected layers with a hidden dimension of $512$. The refinement loss $\mathcal{L}_{ref} $ is formulated as:

\begin{equation}
\label{eq:loss_refine}
    \mathcal{L}_{ref} = \frac{1}{N_S\cdot N_J} \sum_{i}^{N_S} \sum_j^{N_J} \textbf{c}'_{ij} (x_{ij} - \hat{x}_{ij})^2 
\end{equation}

\noindent where we compute the mean-square-error over the number of training samples $N_{S}$ of the predicted poses $\hat{x}$ and normalized ground-truth poses $x$ with joint-wise normalized confidence-weight $\textbf{c}'$.

\begin{figure}[t]
  \centering
  \includegraphics[width=1\linewidth]{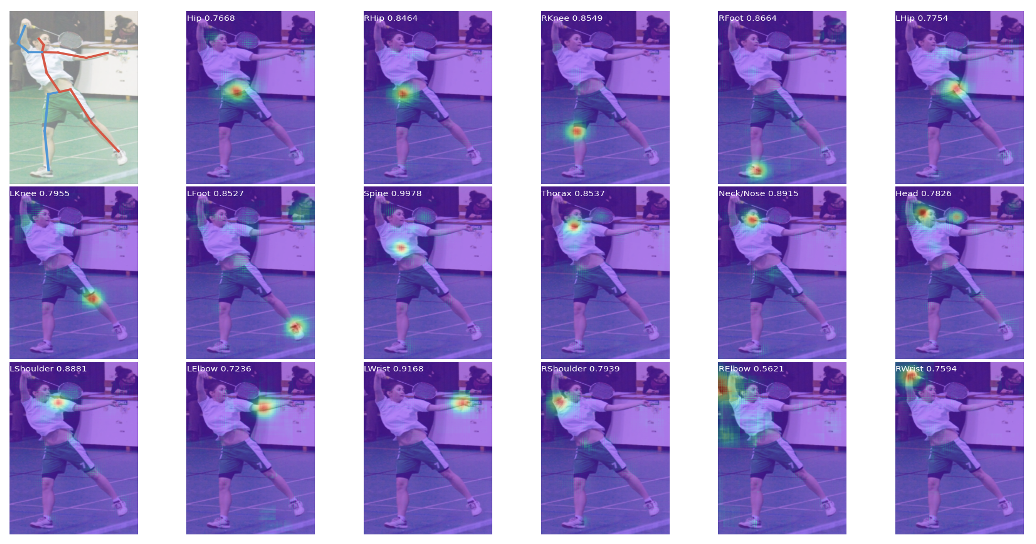}

  \caption{An example of heatmap visualization. Image in the upper left corner is the original image overlaid with the keypoints extracted by HRNet. All the rest images showed the overlaid heatmaps from different keypoints. The maximum scores of each keypoints are different and lower scores indicate lower confidence level.}
  \label{fig:confidence}
\end{figure}
\subsection{Camera Parameter Branch}
In this paper, the 2D-3D pose pairs are calculated in the camera coordinate system, so the camera parameters can be simplified to be the intrinsic matrix $\textbf{M}^{int}$ in Eq.~\ref{eq:intrinsic} and a 3D offset $\textbf{t}_{3D}$. For intrinsic matrix $M_{int}$ we are essentially predicting a $4$-dimensional vector, namely $f_x, f_y, c_x, c_y$, the focal lengths $f_x$, $f_y$, and principal center offsets $c_x$, $c_y$ along the $x$ and $y$ direction respectively. 

\begin{equation}
    \textbf{M}^{int}= \begin{bmatrix}
f_x & 0 & c_x\\
0 & f_y & c_y \\
0 & 0 & 1
\end{bmatrix}
\label{eq:intrinsic}
\end{equation}

\noindent and for the 3D offset $\textbf{t}_{3D}$ we are predicting a $3$-dimensional vector:

%will be our target for the camera parameter branch to predict and reproject the 3D pose back to 2D camera coordinate base via perspective projection \cite{perspective_projection}.
 
\begin{equation}
    \textbf{t}_{3D} = \begin{bmatrix}t_x \\ t_y \\ t_z\end{bmatrix}.
\end{equation}

The camera parameter branch consists of 2 residual blocks with a hidden dimension of $512$, which can be plugged in to any standard 3D pose estimators. There are three losses that can be involved depending on the annotations. The 2D reprojection loss $\mathcal{L}_{2D}$ as shown in Eq.~\ref{eq:2drepro} calculates the Euclidean distance between the reprojected 2D poses and ground truth. The mean-square error (MSE) is used in loss calculation for both the camera parameter loss and 3D inference loss as shown in Eqs.~\ref{eq:camera} and ~\ref{eq:3dpose} respectively.
\begin{equation}
    \mathcal{L}_{2D, \phi'} = \frac{1}{N} \sum_i^{N}\sum_j^{N_j} (\hat{M^{int}}_i \cdot (\hat{\textbf{X}}_{ij} + \hat{\mathbf{t}}_{3D,i}) - x_{ij})^2,
\label{eq:2drepro}
\end{equation}

\begin{equation}
\mathcal{L}_{cam} = ||M^{int} - \hat{M^{int}}||_2^2 + ||\mathbf{t}_{3D} - \hat{\mathbf{t}}_{3D}||_2^2,
\label{eq:camera}
\end{equation}

\begin{equation}
    \mathcal{L}_{3D} = \frac{1}{N_S\cdot N_J} \sum_i^{N_S}\sum_j^{N_j} (\textbf{X}_{ij} - \hat{\textbf{X}}_{ij})^2
\label{eq:3dpose}
\end{equation}

\noindent where $\hat{\textbf{X}}$ stands for the predicted 3D pose from our 3D lifting network.

Since CameraPose can work on 2D-3D pose pairs as well as 2D alone pose estimations, the loss design can be different according to the availability of labels. In the case of all annotations are available during the training stage, the camera loss can be calculated as:
\begin{equation}
    \mathcal{L}_{\boldsymbol\phi} = \lambda_{cam} \mathcal{L}_{cam} + \lambda_{2D, \boldsymbol\phi} \mathcal{L}_{2D, \boldsymbol\phi} + \lambda_{3D} \mathcal{L}_{3D} 
\label{eq:2d3dloss}
\end{equation}

In the case of 2D annotation alone training step, the loss calculation will be from 2D reprojection error:
\begin{equation}
\mathcal{L}_{\boldsymbol\phi'} = \lambda_{2D, \boldsymbol\phi'}\mathcal{L}_{2D, \boldsymbol\phi'}
\label{eq:3dloss}
\end{equation}

\begin{table*}[h]
\centering
\caption{Different human pose estimation datasets used in our work. The datasets in \textbf{bold} font are used for the training while other dataset in \textit{italic} are used for cross-dataset evaluation. The rest of the datasets will be used to visualize and serve as qualitative analysis targets. }
\begin{tabular}{lcccc}
\hline
\multicolumn{1}{c}{Dataset} & \# of Sample & 2D Annotations & 3D Annotations & Camera Parameters \\ \hline
\textbf{Human3.6M}  \cite{h36m_pami}             & 3.6M         & v              & v              & v                 \\ \hline
\textit{MPI-INF-3DHP}   \cite{3dhp}              & 1.3M         & v              & v              & v                 \\ \hline
\textit{3DPW}  \cite{3dpw}                       & 51k          & v              & v              &                   \\ \hline
Ski-Pose  PTZ \cite{SkiPose} & 20k & v & v & v \\ \hline
\textbf{MPII}           \cite{dataset_mpii}              & 25k          & v              &                &                   \\ \hline
MS-COCO  \cite{coco}                    & 250k         & v              &                &                   \\ \hline
\end{tabular}
\label{table:dataset}
\end{table*}
\subsection{Pose Generator and Discriminator}

\begin{figure}[t]
  \centering
  \vspace{-2em}
  \includegraphics[width=1\linewidth]{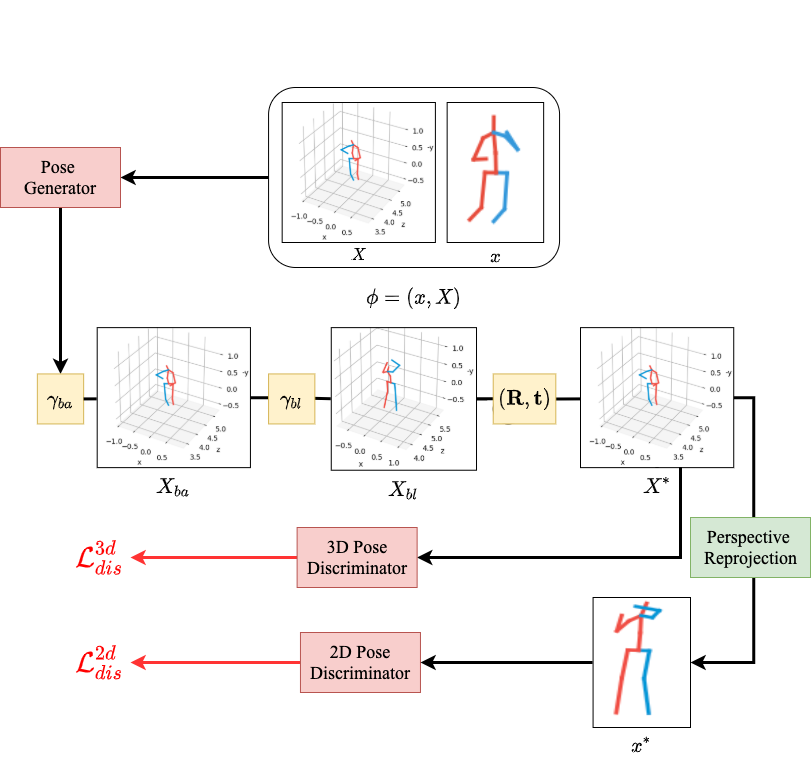}
  \vspace{-1em}
  \caption{Visualization of the pose generator and discriminator. As we augmented from the original 2D-3D annotated dataset $\boldsymbol\phi=(x, \textbf{X})$ using the 3D pose as the input to the generator which give 3 different sets of parameters $\gamma_{ba}$, $\gamma_{bl}$ and $(\textbf{R}, \textbf{t})$ to sequentially modified the 3D pose into our augmented dataset $\boldsymbol\phi^{*}=(x^*, \textbf{X}^*)$.}
  \label{fig:gan}
\end{figure}

Similar to the framework in ~\cite{poseaug}, we utilized both generator and discriminator to further improve the diversity in training poses. As shown in Figure~\ref{fig:gan}, the generator is plugged in to the 2D pose generation stage, and the discriminator is applied on both the 2D and 3D pose inference. 

The generator is actually formed by 3 simple multi-layer perceptions that generated different parameters for 3 different augmentation operations respectively: (1) changing the bone angle $\textbf{X}_{ba}$, (2) changing the bone length $\textbf{X}_{bl}$ and (3) changing the camera view and position of the input 3D pose $\mathbf{R} \cdot \textbf{X}_{bl} + \mathbf{t}$. 

% The discriminators also adapt the part-aware Kinematic Chain Space (KCS) proposed in \cite{poseaug}, they are fully connected networks with a structure similar to the pose regression network using the KCS representation of 2D or 3D poses as input which are first introduced in \cite{repnet}. Here we use the LS-GAN loss:

% A random noise vector is sampled from a Gaussian distribution to concatenate with when feeding the 3D pose $\textbf{X}$ to each of the generators to ensure there is sufficient randomness in the data augmentation framework.

%After obtaining the parameters for each operations, $\mathbf{\gamma_{ba}} \in R^{{(N_{j}-1)}\times 3}$, $\mathbf{\gamma_{bl}} \in R^{{(N_{j}-1)}}$, and $\mathbf{R} \in R^{3\times3}, \mathbf{t} \in R^3$, the original 3D pose $\textbf{X}$ will go through these 3 operations sequentially:

%\begin{align}
%    \textbf &= H_{BA}^{-1}\big(H_{BA}(\textbf{X}) + \mathbf{\gamma_{ba}}\big) \\
%    \textbf{X}_{bl} &= H_{BL}^{-1}\big(H_{BL}(\textbf{X}_{ba}) \times (1 + %\mathbf{\gamma_{bl}})\big)  \\
%    \textbf{X}^* &= \mathbf{R} \cdot \textbf{X}_{bl} + \mathbf{t}
%\end{align}

%\noindent where $H_{BA}$ and $H_{BL}$ is the hierarchical transformation that derived any 3D pose into some bone angle vector and bone length vector. The corresponding 2D pose $x'$ will again being generated by perspective projection using the camera parameters provided in the dataset. Note that in order for such generator to train more smoothly and prevent collapse, we also follow the original work on setting some stricter min-max value for these operations.

The discriminator part of the framework can be divided into 2 portions, the $\mathcal{D}_{2D}$ and $\mathcal{D}_{3D}$ as we want to make sure that both the augmented $\textbf{X}^*$ and $x^*$ formed plausible human poses in both image coordinate and camera coordinate. But in our work we not only want to ensure the goodness of the augmented poses from the generator, we also want to utilized the discriminator to regulated our reprojected 2D poses for those 2D annotations only dataset cases. The discriminators also adapt the part-aware Kinematic Chain Space (KCS) proposed in \cite{poseaug}, they are fully connected networks with a structure similar to the pose regression network using the KCS representation \cite{repnet} of 2D or 3D poses as input. Here we use the LS-GAN loss:

\begin{align}
     \mathcal{L}^{2d}_{dis} &= \frac{1}{2} \mathbb{E}_{x}[(D_{2D}(KCS(x))-1)^2] \\
     &+ \frac{1}{2} \mathbb{E}_{x}[(D_{2D}(KCS( \{ x^*, x'_{2D} \} ))-1)^2] \\
    \mathcal{L}^{3d}_{dis} &= \frac{1}{2} \mathbb{E}_{x}[(D_{3D}(KCS(\textbf{X}))-1)^2] \\
    &+ \frac{1}{2} \mathbb{E}_{x}[(D_{3D}(KCS(\textbf{X}^*))-1)^2]
\end{align}

\noindent as the pose discrimination loss to train the generator and discriminator.

\subsection{Overall Loss}
The overall framework is made differentiable and can be trained in the end-to-end fashion. We update different modules alternatively by minimizing loss in Eq. \ref{eq:loss_refine}, Eq.~\ref{eq:2d3dloss}, Eq. ~\ref{eq:3dloss} as well as generators and discriminators with some preassigned hyper-parameters $\lambda$.

%from the refinement network, generator, 2D and 3D pose discriminator and the pose estimation network with our additional camera branch. However, since our framework consist of multiple sub-networks and even GAN which make it extremely difficult to train all at once. Therefore we still need to first pretrain the 2D keypoints refinement networks and 3D pose estimators using the 2D-3D datasets, Human3.6M individually. And then we can iteratively train the entire model using both types of the datasets: (1) 2D-3D annotated dataset with provided camera parameters and (2) 2D annotations only dataset using the following loss functions:

Then we interactively train the entire model and update the weights of 3D lifting network using the losses:

\begin{equation}
    \mathcal{L}_{\phi} = \lambda_{ref, \phi}\mathcal{L}_{ref} + \lambda_{cam} \mathcal{L}_{cam} + \lambda_{2D, \phi} \mathcal{L}_{2D, \phi} + \lambda_{3D} \mathcal{L}_{3D}
\end{equation}

\noindent and

\begin{equation}
    \mathcal{L}_{\phi'} = \lambda_{ref, \phi'}\mathcal{L}_{ref} + \lambda_{2D, \phi'} \mathcal{L}_{2D, \phi'}.
\end{equation}

\noindent depending on the different datasets $\phi$ or $\phi'$ we are using for the batch. We will introduce more training details and hyper-parameter settings in the Sec. \ref{sec:exp_training}.

\begin{figure*}[t]
  \centering
  \includegraphics[width=1.0\linewidth]{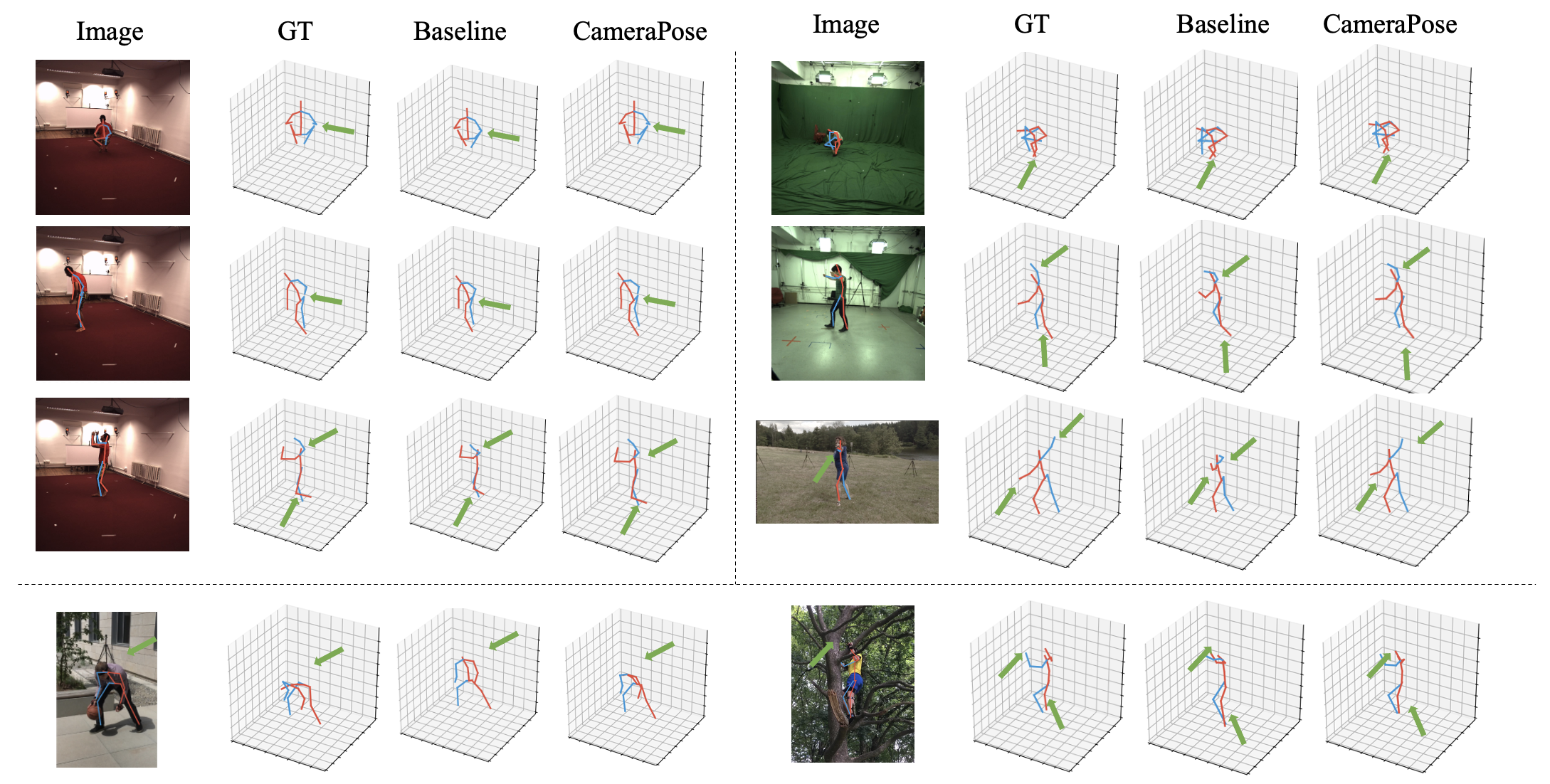}
  \caption{Qualitative comparison for the Human3.6M \cite{h36m_pami} (left), 3DHP \cite{3dhp} (right) and 3DPW \cite{3dpw} (bottom) generalization ability analysis using the pretrained baseline \cite{poseaug} and our purposed method. Both the baseline and our model were trained only with Human3.6M so 3DHP and 3DPW are considered as cross-dataset in this case. The green arrows highlight locations where the models predict differently.}
  \label{fig:compare}
\end{figure*}
\section{Experiments}
\subsection{Datasets}

\begin{table*}[h]
\centering
\caption{Result on Human3.6M, 3DHP and 3DPW using the 2D ground-truth keypoints as the input in terms of MPJPE, note that we use the same model for evaluation on all datasets to mimic cross-dataset evaluation. Best results are shown in \textbf{bold font}. }
\begin{tabular}{lccc}
\hline
\multicolumn{1}{c}{Method}  & Human3.6M (MPJPE) & 3DHP (MPJPE) & 3DPW (PA-MPJPE)  \\ \hline                                                      

Wnadt et al.     \cite{canonpose}            & 74.3                                                        &          104.0                                              &       -                                                    \\

Rhodin et al.     \cite{SkiPose}            & 80.1                                                        &          121.8                                              &       -                                                    \\ \hline

Zhao et al.     \cite{GCN}            & 44.4                                                        &          97.4                                              &       -                                                    \\
Martinez et al.         \cite{martinez_2017_3dbaseline}           & 43.3                                                        &                 85.3                                      &      -                                                     \\
Cai et al.       \cite{stgcn}          & 41.7                                                        &           87.8                                             &   -                                                        \\
Pavllo et al.       \cite{videopose}       & 41.80                                                        & 92.64                                                        &    76.38                                                       \\
Gong et al.     \cite{poseaug}          & 39.02                                                       & \textbf{76.13}                                                  & 66.27                                                     \\ \hline
\textbf{Ours   (CameraPose)}                    & \textbf{38.87}                                                       & 78.85                                                  & \textbf{63.26}                                                     \\ \hline
\end{tabular}
\label{table:overall}
\end{table*}

For the 2D-3D paired annotations, we utilize the most popular datasets 3D HPE dataset \textbf{Human3.6M} \cite{h36m_pami}, \textbf{3DHP} \cite{3dhp} and \textbf{3DPW} \cite{3dpw}. Both Human3.6M and 3DHP were collected indoor in some laboratory environment through the MoCap (motion capture) system \cite{mocap} with multiple calibrated cameras. The 3DPW is a more challenging dataset collected in outdoor environment using IMU (inertial measurement unit) sensors with mobile phone lens.

For 2D annotations only datasets, we used \textbf{MPII} \cite{dataset_mpii} which contains a variety of in-the-wild everyday human activities. Another popular 2D dataset \textbf{MS-COCO} \cite{coco} is also used for qualitative analysis purposes.  Although the 2D annotation dataset such as MPII is much less than Human3.6M or 3DHP in terms of sample size, these 2D annotation datasets contain more challenging human poses with different activities. Note that both Human3.6M and 3DHP are video based datasets, so that the total number of images is much larger than MPII and MSCOCO. We summarized the datasets utilized in our experiments in Table \ref{table:dataset}.

\subsection{Preprocessing}
Different datasets have distinctive annotations on joints, which make the model training difficult. In this paper, we used the Human3.6M format as standard one, and interpreted missing joints by labeling nearby joints for other datasets. All the joints that are not included in Human3.6M format will be discarded. 

Many existing 3D HPE algorithms use the groundtruth as model input for evaluation. However, groundtruth is not available in real use cases. To evaluate the model performance on the real-world applications, we also used existing 2D detector HRNet to extract the 2D keypoints as model input and rerun results on different datasets.

%Due to the various labeling schemes or joint formats difference (shown in Fig. \ref{fig_preprocess}), 

Due to the various labeling schemes or joint formats difference, we preprocess other schemes into the Human3.6M format by simple interpolation of some related joints and removal of the unused joints. For example, there is no pelvis; we simply create such joint by computing the mid-point of the left and right hip of any given label. Even though such interpolations are not always perfect due to the nature of each dataset, this preprocessing procedure allows us to have a better idea and comparison on cross-dataset scenarios.

\subsection{Training} \label{sec:exp_training}
CameraPose network is trained on 2 datasets: Human3.6M (2D + 3D) and MPII (2D). For the former, we followed most 3D human pose estimation training protocols using the subjects S1, S5, S6, S7, S8 from Human3.6M as our 2D-3D training data, and subjects S9, S11 for evaluation purposes. For the latter, we filtered and selected around 10k training samples by checking the joints annotations. For evaluation, MPI-INF-3DHP and 3DPW were used to get quantitative results in terms of MPJPE (mean-per-joint-position-error) and PA-MPJPE (aligned with ground-truths by rigid transformation).

The model training can be divided into 3 steps. The refinement network was trained as the first step for 100 epochs with learning rate being 0.0001 and weight decay at epochs 30, 60 and 90, respectively. Next step, the 3D lifting network along with the pose generator and discriminator was trained using Human3.6M dataset for 10 epochs with a learning rate of 0.0001. This step is for warm-up and GAN tuning which can make the following model training more stable. Finally, the model was trained in an end-to-end fashion using both 2D-3D pairs annotations as well as 2D alone annotations. In each iteration, we first updated the weights of generator and discriminator to make the generators more stable. Then the 3D lifting network was updated based on the augmented poses plus the 2D-3D annotated dataset. After that, 2D only annotations were utilized to tune the camera parameter branch. The model was trained for 75 epochs with a learning rate of 0.0005 and weight decay at 30, 60, respectively. And the weighting for loss we choose $\lambda_{cam} = 0.01$, $\lambda_{2D, \boldsymbol\phi} = 0.5$, $\lambda_{2D, \boldsymbol\phi'}=0.2$, and $\lambda_{3D} = 1.0$.

\begin{table*}[t]
\caption{Experimental results of the effect of refinement network as we examine the effectiveness of our refinement module using the HRNet detections on training and evaluation purposes.}
\centering
\label{tab:my-table}
\begin{tabular}{lccc}
\hline
\multicolumn{1}{c}{Method} &
  \begin{tabular}[c]{@{}c@{}}Training Source\\ (2D Estimator)\end{tabular} &
  \begin{tabular}[c]{@{}c@{}}Human3.6M\\ (MPJPE)\end{tabular} &
  \begin{tabular}[c]{@{}c@{}}3DHP\\ (MPJPE)\end{tabular} \\ \hline
Pavllo et al.    \cite{videopose}                 & Human3.6M (HRNet)         & 57.90 & 103.86 \\ 
Gong et al. \cite{poseaug} & Human3.6M (HRNet)         & 55.18 & 99.50  \\
\hline
Gong et al. \cite{poseaug} w/ \textbf{Refinement Network}                    & Human3.6M (HRNet)       & 54.32 & 97.45  \\ \hline
\textbf{CameraPose w/ Refinement Network} & Human3.6M (HRNet) & \textbf{54.20} & \textbf{97.35}  \\
\textbf{CameraPose w/o Refinement Network} & Human3.6M (HRNet) & 54.38 & 98.12  \\

\hline
\end{tabular}
\end{table*}

\subsection{Quantitative Results}
\noindent \textbf{CameraPose Network Accuracy}. We compared CameraPose with other state-of-the-art methods \cite{stgcn,videopose,GCN,poseaug,martinez_2017_3dbaseline} trained on Human3.6M. For the temporal-based methods \cite{stgcn,videopose}, we implemented the single frame version for a fair comparison. Table ~\ref{table:overall} summarized the experimental results of different methods. For each column, the MPJPE or PA-MPJPE are calculated for evaluation, obtained from the same model trained and selected based on the evaluation dataset of Human3.6M. Some existing algorithms selected distinctive best models on different testing datasets, which may not reflect the generalization of models well. Instead, we selected a single model based on the accuracy of the validation of Human3.6M to make it more realistic for real-world application.

\begin{figure}[t]
  \centering
  \includegraphics[width=1\linewidth]{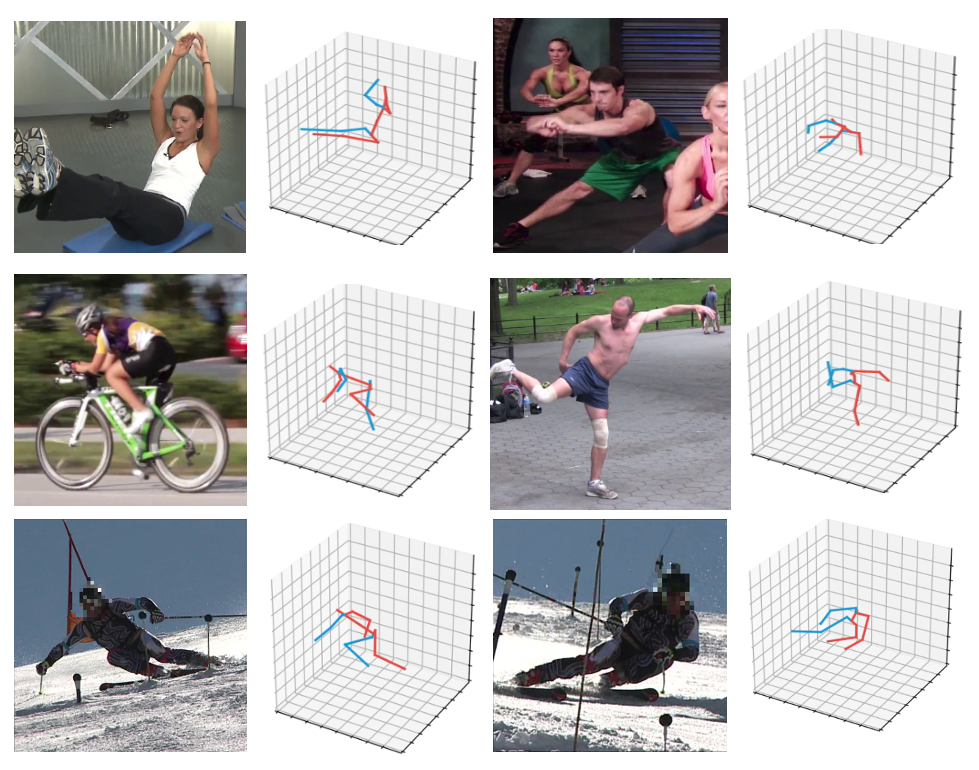}

  \caption{Visualization for qualitative analysis of 3D human pose estimation on MPII \cite{dataset_mpii} (Testing), MS-COCO \cite{coco}, and SKiPose-PTZ \cite{SkiPose}. Our model can still generate reliable 3D poses even when the target poses are in general rare or never seen from the training.}
  \label{fig:morevis}
\end{figure}

As shown in Table \ref{table:overall}, our method outperforms the SOTA on the most challenging dataset 3DPW by a noticeable margin ($3$mm and $13$mm). It also has significantly higher accuracy than other weakly-supervised methods like \cite{SkiPose} and \cite{canonpose}. Our model also achieves the highest accuracy on the Human3.6M dataset. Experimental results clearly show the strong generalization capability of our proposed method. Adding the camera parameter branch can help the model to learn from in-the-wild datasets with 2D annotations, which is very effective for hard examples. 

The results on the 3DHP are slightly lower than SOTA methods, and we claim it is due to the fact that the 2D annotations we added from MPII are more helpful for challenging cases such as the 3DPW dataset. The best accuracy on the 3DHP dataset can be 75.54 MPJPE using our model, which outperforms the current SOTA if we select a specific model for the 3DHP dataset. 

\noindent \textbf{Refinement Network Accuracy}. To show the effectiveness of the refinement network, we trained different models with different settings as shown in the Table ~\ref{tab:my-table}. 

We used HRNet as 2D detectors to extract the 2D keypoints on all the training and evaluation datasets. We added the refinement network to both the SOTA method~\cite{poseaug} and our proposed model. By adding the refinement network, both PoseAug and our model have improved accuracy on both Human3.6M and 3DHP. In addition, our model outperforms the SOTA on both testing datasets. Therefore, both our proposed camera parameter network and the refinement network are useful for 3D HPE.

\begin{figure}[t]
  \centering
  \includegraphics[width=0.95\linewidth]{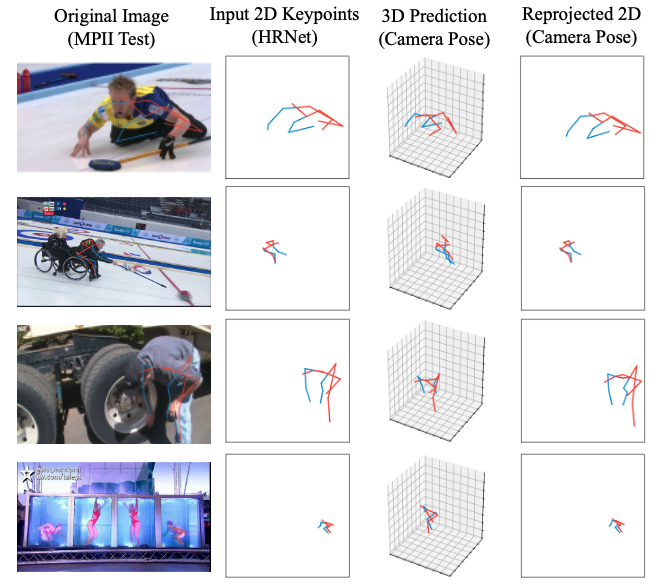}

  \caption{3D-2D reprojection visualization on MPII \cite{dataset_mpii}. Column from the left: original images, 2D keypoints from HRNet, inferred 3D keypoints, reprojected 2D keypoints. The camera parameters predicted by CameraPose can successfully reprojected the 3D pose back into the image coordinate.}
  \label{fig:reprojection}
\end{figure}

\subsection{Qualitative Visualization}
\noindent \textbf{3D Pose Estimation}. We choose 3 datasets (Human3.6M, 3DHP and 3DPW) to qualitative compare our proposed method and baseline~\cite{poseaug}. As shown in Figure ~\ref{fig:compare}, our model has more accurate predictions on challenging datasets such as 3DPW. Note that we utilize cross-scenario training to make sure there is no overlap between training and testing datasets. We also visualize our results on datasets without 3D annotations such as MPII, MSCOCO, and SkiPose-PTZ~\cite{SkiPose} in Figure ~\ref{fig:morevis}. The visualization results are very plausible, which indicates the capability of our model for in-the-wild prediction.

\noindent \textbf{2D Reprojection}. To validate the camera parameter branch, we visualize the results of our model at a different stage. Figure ~\ref{fig:reprojection} shows the original image, input 2D keypoints from HRNet, inferred 3D poses, and reprojected 2D poses from left to right columns. It clearly shows that our CameraPose can predict well on unseen poses and the reprojected 2D poses are meaningful too.

%\subsection{Ablation Studies}

% To better reproduce and compare the result on other works as well \cite{GCN, videopose, poseaug}, we use our own HRNet 2D detection results and its corresponding confidence scores as input. In Table \ref{table:gt}, we try to analyze whether cleaner 2D keypoints can really benefit the 3D lifting model by using different sources of 2D keypoints with the different 3D pose estimator. As you can see, the ground-truth yield a very large margin in compare to the 2D detectors, which encourage us to impose the refinement network first before feeding the 2D keypoints into the 3D lifting network. 

\section{Conclusions}

We propose CameraPose, a weakly-supervised framework for 3D human pose estimation from a single image that can aggregate 2D annotations by designing a camera parameter branch. Given any noisy 2D keypoints from pretrained 2D pose estimator, CameraPose is able to refine the keypoints with a confidence-guided loss and feed them into the 3D lifting network. Since our approach uses the camera parameters learned from the camera branch to do the reprojection back to 2D, it can solve the problem of the lacking of the 2D-3D datasets with rare poses or outdoor scenes. We evaluate our proposed method on some benchmark datasets; the results show that our model can achieve higher accuracy on challenging datasets and be able to predict meaningful 3D poses given in-the-wild images or 2D keypoints.  

{\small
\bibliographystyle{ieee_fullname}
\bibliography{egbib}
}

\end{document}